\documentclass[letterpaper, 10 pt, conference]{ieeeconf}  

\IEEEoverridecommandlockouts                              

\usepackage{times}
\usepackage{epsfig}
\usepackage{graphicx}
\usepackage{amsmath}
\usepackage{amssymb}

\usepackage{siunitx}
\usepackage{longtable,tabularx}
\usepackage{multirow}
\usepackage{textcomp}
\usepackage{gensymb}
\usepackage{array}
\usepackage{verbatim}
\usepackage{comment}

\title{\LARGE \bf
Multi-layer VI-GNSS Global Positioning Framework with Numerical Solution aided MAP Initialization
}

\author{Bing Han$^{1}$, Zhongyang Xiao$^{1}$, Shuai Huang$^{1}$ and Tao Zhang$^{1,*}$

\thanks{$^{1}$Bing Han, Zhongyang Xiao, Shuai Huang and Tao Zhang are with Alibaba Group, Beijing, China,
{\tt\small nikchyhahn@aliyun.com},
{\tt\small young.xzy@autonavi.com},
{\tt\small \{ hanzhe.hs, zt107579\} @alibaba-inc.com}} %
\thanks{* corresponding author}
}

\begin{document}

\maketitle
\thispagestyle{empty}
\pagestyle{empty}

\begin{abstract}

Motivated by the goal of achieving long-term drift-free camera pose estimation in complex scenarios,  we propose a global positioning framework fusing  visual, inertial and Global Navigation Satellite System (GNSS) measurements in multiple layers. Different from previous loosely- and tightly- coupled methods, the proposed multi-layer fusion allows us to delicately correct the drift of visual odometry and keep reliable positioning while GNSS degrades. In particular, 
local motion estimation is conducted in the inner-layer, solving the problem of scale drift and inaccurate bias estimation in visual odometry by fusing the velocity of GNSS, pre-integration of Inertial Measurement Unit (IMU) and camera measurement in a tightly-coupled way. The global localization is achieved in the outer-layer, where the local motion is further fused with GNSS position and course in a long-term period in a loosely-coupled way. Furthermore, a dedicated initialization method  is proposed to guarantee fast and accurate estimation for all state variables and parameters. We give exhaustive tests of the proposed framework on indoor and outdoor public datasets. The mean localization error is reduced up to 63\%, with a promotion of 69\% in initialization accuracy compared with state-of-the-art works. We have applied the algorithm to Augmented Reality (AR) navigation, crowd sourcing high-precision map update and other large-scale applications.

\end{abstract}

\section{INTRODUCTION}

Real-time estimation of 6-DOF camera pose has become a fundamental requirement in many automated systems such as robotics, autonomous vehicle navigation, and AR.
Advanced solutions \cite{Davison2007MonoSLAM,qin2018vins,lategahn2011visual} 
simultaneously estimated the camera pose while constructing the environment map  by detecting and tracking visual features, which is named as simultaneous localization and mapping (SLAM).
Among SLAM approaches, 
monocular SLAM has the advantages of low cost, small size and easy installation. However, it suffers from inherent problems of scale uncertainty and scale drift.

Though stereo \cite{Mur2017ORB} and RGB-D \cite{fu2019robust} camera systems can solve the problem of scale uncertainty, the cost of hardware and computation complexities are significant.
Recently developed visual-inertial (VI) systems \cite{qin2018vins,huang2018online} solve the problem at a low cost. On one hand, IMU measurement helps recover the metric scale. On the other hand, the IMU bias can be corrected by vision information.

Due to the cumulative error in both visual and inertial systems, however, the above-mentioned VI systems can guarantee positioning accuracy only in a short time period, which limits its application in long-term and large-scale area. Loop closure can reduce the positioning drift, but will increase the computational cost. Moreover, large-scale outdoor applications can hardly benefit from loop closure \cite{zhang2017loop,naseer2015robust}. 
Another drawback in VI systems is that the estimated pose is not aligned to the world frame, which is essential for applications (e.g, the intelligent vehicle) using a priori global map.
Among other commonly used localization methods,
GNSS provides global positioning information in the world frame with time-independent accuracy. 
It can help eliminate the cumulative error and correct the scale drift.
Visual, inertial and GNSS (VI-GNSS) integration forms a promising framework for outdoor long-term positioning solution  \cite{shin2020dvl,surber2017robust,yu2019gps,niesen2016robust}.

In this paper, we propose an innovative VI-GNSS global positioning framework, which can realize rapid initialization and online localization in the world frame.  Even in challenging scenarios, it guarantees reliable positioning performance with robustness to GNSS degradation and visual instability.

\begin{figure}
    \centering
    \includegraphics[width = 0.35 \textwidth]{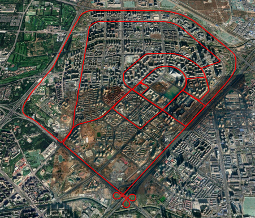}
    \caption{
    Global localization result of the proposed method
    }
    \label{wangjing_result}
\end{figure}

The main \textbf{contributions} of this paper are as follows:

1. We propose a novel Maximum a Posteriori (MAP)-based VI-GNSS joint initialization framework. Compared with other visual initialization algorithms, it has the advantages of shorter initialization time, more accurate scale estimation, and online estimation of the extrinsic parameters between the local and world frames.

2. We propose an innovative multi-layer fusion framework, in which the translation and rotation estimation is dedicatedly decoupled according to the observation properties from different sources to achieve optimized estimation in each layer.

3. Exhaustive experiments and applications 
are conducted to show that, compared with the existing algorithms, our proposed method can achieve 
 higher accuracy 
up to 69\% in the initialization and 63\% in the online localization in large-scale challenging scenarios.

\section{RELATED WORKS}

In recent decades, there have been many efforts to realize  camera pose estimation by fusing visual, inertial sesnors and GNSS, wheel encoder, etc .
Among them, Visual SLAM is the basis for fusing other sensors.

\textbf{Visual SLAM}
The mainstream visual SLAM methods include indirect methods \cite{klein2007parallel,Mur2017ORB,Davison2007MonoSLAM} and direct methods \cite{engel2013semi,forster2014svo}. 
Since the depth of visual feature is ambiguous, the monocular visual SLAM can provide only up-to-scale poses. Some solutions use a 
binocular or RGB-D camera \cite{hu2012robust} or lidar \cite{shin2018direct} to determine the depth directly. Other methods that introduce inertial sensor to restore the real scale \cite{bloesch2015robust,Tong2018VINS} gain more interest due to its relatively low cost. However, when the acceleration excitation is low (e.g., the vehicle moves at a constant speed), the metric scale is unobservable and easily diverged.

Visual SLAM can estimate the pose only relative to the first camera frame.
However, many applications (e.g., the intelligent vehicles, etc.) require the global pose in the world frame so that it can be aligned to a prior map.
Therefore, multi-source sensors should be fused to obtain the global pose.
According to the different frameworks where multiple sensors are fused, the fusion algorithms 
are usually divided into loosely-coupled positioning and tightly-coupled positioning.

\textbf{Loosely-coupled positioning v.s. Tightly-coupled positioning}.
%
In the framework of loosely-coupled fusion 
\cite{qin2019general,mascaro2018gomsf,surber2017robust}
, parts of the original sensor data (e.g., visual and inertial measurements) are fused at the first stage and produce the estimation as a new measurement.
The pre-estimated data is further fused with other sensors. The process of modeling errors in the pre-esitmation may cause inaccuracy.
Moreover, the drift of the metric scale in visual SLAM cannot be corrected for a long time, which may eventually cause a large error in the final positioning result.
The limitations can be overcome by
directly modeling the noise of various sensors' observations and fusing them in one single sliding window. The improvement leads to the tightly-coupled positioning that guarantees higher accuracy \cite{cioffi2020tightly,yu2019gps}. However, the  tightly-coupled system is more nonlinear and has a higher degree of freedom. Abnormal data from single sensor may cause large drifts in the final positioning.

The proposed
framework in this paper is between loosely- and tightly- coupled methods: we firstly fuse a portion of data from visual, inertial and GNSS in the inner-layer in a small time period using a tightly-coupled way. In addition to accurately modeling the observation noise, we deal with the visual-inertial scale drift by introducing GNSS velocity. Then we fuse the GNSS position and course in a longer time period in the outer-layer to further improve the global pose accuracy. Also, outliers are eliminated in this layer, making it more robust in scenarios with large GNSS drift and short-term visual measurement degradation.

\textbf{Parameter initialization methods}
In the mainstream VI SLAM algorithms \cite{qin2018vins,Mur2017ORB,mur2017visual}, initial state variables (e.g., camera pose, metric scale, etc. ) are obtained by solving a numerical analytic solution,
which are not accurate enough.
In the latest research Orb-slam3 \cite{campos2020inertial},  MAP method is used to estimate the initial state variables in an optimization framework, which is prone to achieve more accurate initialization. However, Orb-slam3 does not finely set the initial values in MAP optimization, and may result in a local optimum result. 
Our proposed algorithm roughly estimates initial values using the analytical method and then performs further optimization by MAP estimation.
In addition, 
we involve the initialization of local-global transformation to guarantee accurate estimation of global poses.

\section{INITIALIZATION}

In the MAP based monocular VI-GNSS online optimization system, a pure monocular camera can provide relative pose information where the scale is normalized. IMU helps recover the real metric scale, however the accuracy suffers from the drift of bias. GNSS provides the position, velocity and course information in the world frame(i.e., East-North-Up (ENU) navigation coordinate system) with relatively low accuracy. 
The initialization procedure should guarantee good initial values for all state variables and parameters, e.g. the metric scale, IMU bias, extrinsic parameters of the world frame, etc., to avoid diverging or converging to a local minimal.



In this section, we propose a numerical solution aided MAP initialization algorithm where 
initial values are calculated in a uniform framework.
It can also be applied to other VI GNSS and/or odometer systems.


The entire initialization algorithm includes 3 steps:


\begin{enumerate}
\item Estimate the initial pose with real metric scale using sliding window visual odometry (VO) aided by GNSS velocity.
\item Solve the extrinsic parameters of the initial camera frame against the world frame.
\item Estimate and refine all parameters using the VI-GNSS MAP optimization.
\end{enumerate}


\begin{figure*}
    \centering
    \includegraphics[scale = 0.45]{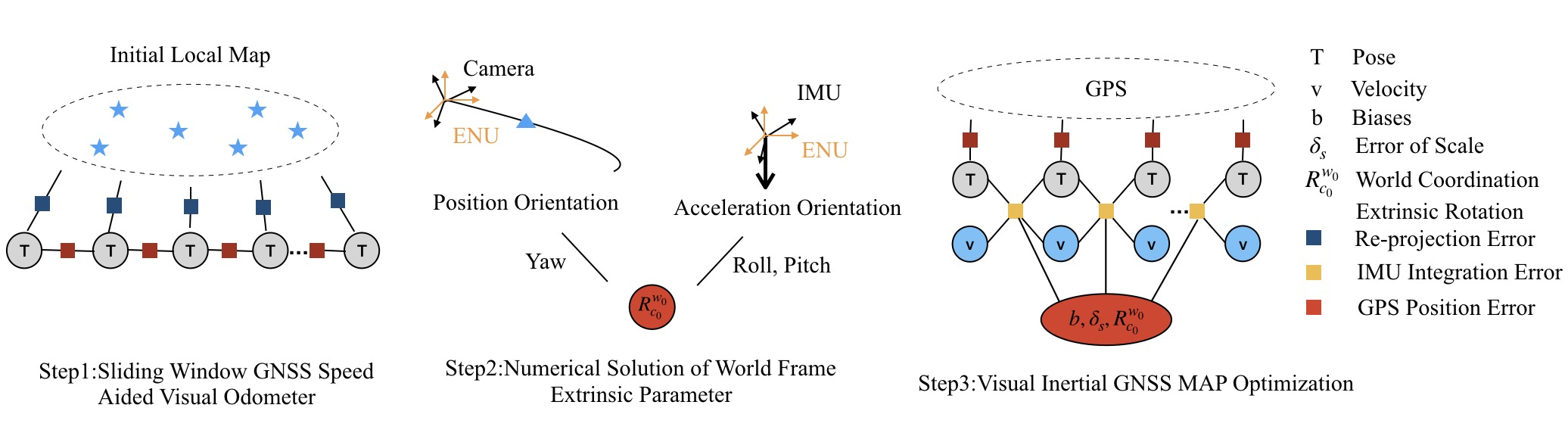}
    \caption{ Initialization Procedure }
    \label{figure_init}
\end{figure*}
   
\subsection{Sliding window VO aided by GNSS velocity}

In this step, we roughly estimate the initial pose with real metric scale while constructing the initial 3D feature point map simultaneously. In order to guarantee the real-time calculation performance, we use a fixed-length sliding window to maintain the data from latest several frames. For all the matched frames, when the number of tracking points and the parallax are sufficient and the GNSS signals are good , we calculate the relative rotation $R_0$ and translation $t_0$ by the 7-point method \cite{hartley2003multiple}. Furthermore, we retrieve the metric scale by comparing the relative translation of the two matched frames against the GNSS velocity. The initial 3D visual point map is then constructed by triangulating all matched feature points. Poses of other frames are estimated by perspective-n-point (PnP) method \cite{lepetit2009epnp}. Finally, we take a GNSS-aided visual bundle adjustment (BA) in Equation \ref{ba_initial} to optimize poses of all frames in the sliding window.





 \begin{align}
     \begin{split}
         &X = [\delta s,p^{c_0}_{c_0},q^{c_0}_{c_0},p^{c_0}_{c_1},q^{c_0}_{c_1}...p^{c_0}_{c_k},q^{c_0}_{c_k},{\lambda}_0,{\lambda}_1...{\lambda}_m] \\
          &X^* = \arg \min_X \{ {  \rho ( \lVert r(z_{c_k},X) \rVert}^2) + 
{\lVert r(z_{\Delta p},X) \rVert} ^ 2
 \} \\
  &r(z_{c_i}, \chi)= {q_{c_j}^{c_0}}^{-1} \otimes 
  \left[ {q_{c_i}^{c_0}} * p_i / {\lambda}_l + p^{c_0}_{c_i} -  p^{c_0}_{c_j} \right] - p_j
  \\
   &r(z_{\Delta p},X) = \frac{1}{2} * (v_i + v_j) * \Delta t -  s * \lVert p^{c_0}_{c_j} - p^{c_0}_{c_i} \rVert
     \end{split}
     \label{ba_initial}
 \end{align}
 
 where $r(z_{c_k},X)$ is the visual re-projection error.
 $r(z_{\Delta p},X)$ is the relative translation error of corresponding frames.
 $ p^{c_0}_{c_i},q^{c_0}_{c_i} $ is the pose of frame i relative to initial frame.
 $\rho(x)$ is Huber robust loss function which is used to reduce the influence of outliers \cite{barron2019general}.
 There are few outliers in the translation
error estimates, so $\rho(x)$ is not applied to relative translation error.
 ${\lambda}_l$ is the inverse depth of feature point in the firstly observed camera frame. $p_i$ is the position of feature in the normalized plane.
 $v_i,v_j$ are the velocities of GNSS at the time of i-th and j-th
 frames, and $\Delta t$ the time interval between the i-th and j-th
 frames. 
 We estimate the metric scale using the relative translation derived from GNSS velocities which are considered more accurate than GNSS positions thanks to the Doppler measurement.




\subsection{Numerical solution for world frame extrinsic parameter}

The poses acquired in the former step are relative to the initial camera frame, however the global positions are required to support outdoor large-scale applications. So it is essential to estimate the relative rotation extrinsic parameters between the initial camera frame and the world frame.

To estimate the 3-DOF rotation extrinsic parameters, the system should render the roll, pitch, and yaw angles observable. Firstly, we align the GNSS positions in the world frame with the camera positions in the first frame to derive the initial yaw angle.



\begin{equation}
    p^w_{g_k} = q^w_{c_0}(y) \otimes \{  s * p^{c_0}_{c_k} - q^{c_0}_{c_k}\otimes {q^b_c}^{-1} \otimes (t^b_g - t^b_c)\} 
\end{equation}

where 
$q^w_{c_0}(y)$ 
is the yaw angle in the extrinsic parameters.
$q^b_c,t^b_c$ 
are the rotation and translation parameters between camera and IMU frame.
$t^b_g$
is the translation between IMU and GNSS antenna in the IMU body frame.


Then, we estimate roll and pitch angles using the acceleration of IMU, where the motion acceleration is eliminated by differentiating the GNSS velocity.


\begin{equation}
    q^w_{c_0}(r, p) \otimes q^{c_0}_{c_k} \otimes {q^i_c}^{-1} * a_i 
= (G + a_l)
\end{equation}

where 
$q^w_{c_0}(r, p)$ 
are roll and pitch angles of extrinsic parameters.
$a_i$
is the acceleration measurement from IMU.
$a_l$
is the motion acceleration in the world frame.


Finally, we combine $q^w_{c_0}(r, p)$, $q^w_{c_0}(y)$ to form $q^ w_{c_0}(r, p, y)$ as the initial rotation extrinsic parameters between the world frame and the initial camera frame.


\subsection{VI-GNSS MAP optimization}

In the last two steps, we have roughly estimated the initial metric scale by GNSS and VO fusion, however it can be further optimized using the IMU measurements. Besides, the rotation extrinsic parameters are only estimated by single or a few frames and is not accurate enough. We use the roughly estimated parameters as the first guess and conduct a MAP optimization using all information in the first few frames to obtain more accurate initial values.          


\begin{equation}
    \begin{split}
        &X_0 = \left\{ b_g,  s, q^w_{c_0}, v_0, v_1, v_2 ... v_n \right\}\\
    &X_0^* = argmin_X \left\{ 
{ \lVert r(z_{p},X) \rVert} ^ 2 _{ \Omega_g} + 
{\lVert  r(z_{b_k}^{b_{k+1}},X) \rVert} ^ 2
 \right\} \\
 &r(z_{p},X) = q^w_{c_0} \otimes \{  s * p^{c_0}_{c_k} - q^{c_0}_{c_k}\otimes {q^i_c}^{-1} \otimes (t^i_g - t^i_c)\} 
- p^w_{g_k}
\\
&r(z_{b_k}^{b_{k+1}},X) = \\
&\left(
\begin{array}{c}
     \alpha_{b_k}^{b_{k+1}} - 
{q^{c_0}_{b_k}}^{-1} \otimes {q^w_{c_0}}^{-1} \otimes
(p^w_{b_{k+1}} - p^w_{b_k} +\\ \frac{1}{2} G \Delta t ^ 2 - v^w_{b_k}\Delta t)  \\
\beta_{b_k}^{b_{k+1}} -
{q^{c_0}_{b_k}}^{-1} \otimes {q^w_{c_0}}^{-1} \otimes
(v^w_{b_{k+1}} - v^w_{b_k} + G \Delta t)\\
{q^{c_0}_{b_{k+1}}}^{-1} \otimes
q^{c_0}_{b_{k}} \otimes
\gamma^{b_{k}}_{b_{k+1}} \otimes
\dbinom{1}{\frac{1}{2}*J^{\gamma}_{b_g}\delta {b_g}}
^2
\end{array}
\right)
    \end{split}
\end{equation}

where
$r(z_{p},X)$ is the translation residual and
$r(z_{b_k}^{b_{k+1}},X)$ the pre-integration residual \cite{qin2018vins}. $b_g$ is the bias of gyroscope. $\delta s$ is the scale error. $q^w_{c_0}$ is the extrinsic parameter between the initial camera frame and the world frame. $v_i$ is the velocity of IMU relative to the world frame.
We can solve this MAP problem with roughly estimated parameters as initial values using Levenberge-Marquart algorithm to obtain more accurate states. At each iteration, we substitute the latest $b_g,  s, q^w_{c_0}$ in case of getting into local minimal.

\section{ONLINE POSE ESTIMATOR}

\begin{figure*}
    \centering
    \includegraphics[width = \textwidth]{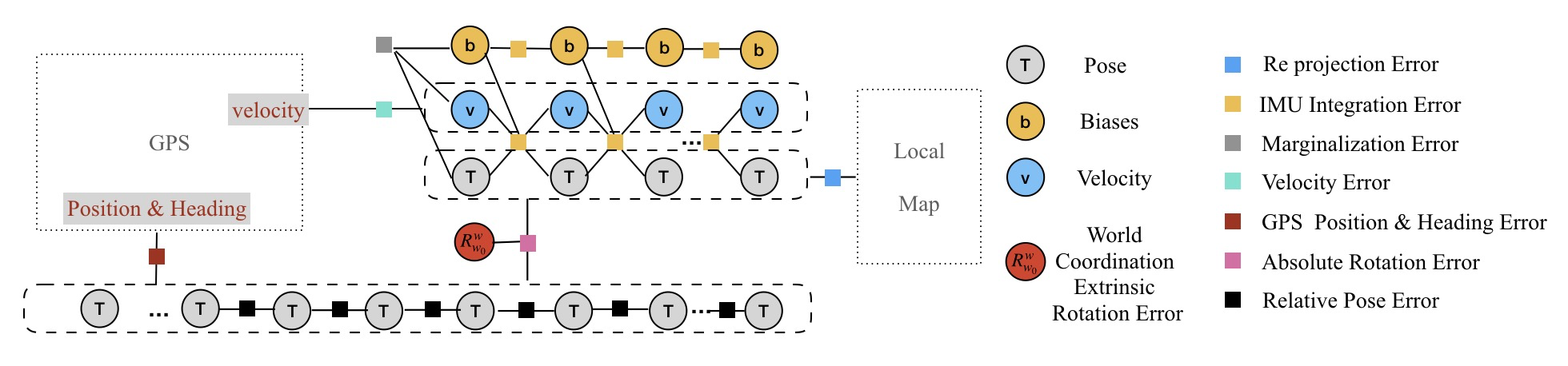}
    \caption{ Factor Graph of Multi-layer VI-GNSS Global Positioning Framework}
    \label{factorGraph.png}
\end{figure*}

The online estimation process is shown in Figure \ref{factorGraph.png}. This framework consists of two layers. 1) The inner-layer is the sliding window BA optimization. And, 2) the outer-layer is 4-DOF pose graph optimization in a large scale.


\subsection{Sliding window BA}
In the sliding window BA, we mainly fuse visual inertial measurements and GNSS velocity to get accurate poses, which can simultaneously calibrate the extrinsic parameters between camera and IMU body frames. The states to be estimated in the sliding window are


\begin{align}
    \begin{split}
        &\chi = \left[\xi_0, \xi_1, \xi_2 ... \xi_m, T^b_c, \rho_0, \rho_1 ... \rho_n \right]\\
        &\xi_k = \left[p^{w_0}_{b_k}, q^{w_0}_{b_k}, v^{w_0}_{b_k}, b_\omega, b_a\right]\\
        &T^b_c = \left[p^{b}_{c}, q^{b}_{c}\right]
    \end{split}
\end{align}

where $\xi_k$ is the state of IMU at k-th frame.
$p^{w_0}_{b_k}$, $q^{w_0}_{b_k}$, $v^{w_0}_{b_k}$ are the pose and velocities of IMU at k-th frame.
$b_\omega$, $b_a$ are the biases of gyroscope and accelerometer respectively.
$\rho_k$ is the inverse depth of visual feature.
m is the number of frames in the sliding window and n the number of features in the visual feature point map.


By optimizing the residuals of multiple sensors in the sliding window , we can obtain accurate  poses and camera-IMU extrinsic parameters. A marginal method is required to avoid the loss of information caused by the direct removal of historical frames. The Mahalanobis norm is used to normalize the observation residuals. The overall cost function is


\begin{equation}
    \begin{split}
      \chi^* = arg min_{\chi} \left\{
\left \| r_m - H_m * \chi \right \|^2 
+
\sum_{i=0}^n \rho ( \left \| r(z^{c_i}, \chi ) \right \|^2_{\Omega_{c_i}}) 
\right. \\ \left.
+
\sum_{j=0}^{m-1} \left \| r(z^{b_{j+1}}_{b_j}, \chi ) \right \|^2_{\Omega_{b_j}} 
+ 
\sum_{k=0}^{m} \rho (\left \| r(z^{g_{v_k}}, \chi ) \right \|^2_{\Omega_{g_{v_k}}}) 
\right\}  
    \end{split}
\end{equation}

where
$H_m$ is the information matrix of a priori residuals.
$r(z^{c_i})$ is the re-projection residual.
$r(z^{b_{j+1}}_{b_j})$ is the IMU pre-integration residual.
$r(z^{g_{v_k}})$ is the velocity residual.
$\Omega_{c_i},\Omega_{b_j},\Omega_{g_{v_k}}$ are Hessian matrixes of re-projection, IMU pre-integration and velocity.

The residual of re-projection is

\begin{equation}
    r(z^{c_i}, \chi)= {q_c^b}^{-1} \left[ {q_{b_i}^{w_0}}^{-1} *
(X^{w_0}_i - p_{b_i}^{w_0}) - t_c^b \right] - p_{c_i}
\end{equation}

where
$p_{c_i}$ is the normalized position of visual feature in the camera frame.
$X^{w_0}_i$ is the 3D position of visual feature point in the initial world frame. 

The residual for IMU pre-integration is

\begin{align}
    \begin{split}
       & r(z^{b_{j+1}}_{b_j}, \chi) = [r_q, r_p, r_v, r_{b_\omega},  r_{b_a}]\\
       & r_q = {\Delta q_{b_k}^{b_{k+1}}}^{-1} \otimes {q^{w_0}_{b_{k}}}^{-1} \otimes q^{w_0}_{b_{k+1}}\\
       & r_p = {q^{w_0}_{b_{k}}}^{-1} \otimes 
\left( p^{w_0}_{b_{k+1}} - p^{w_0}_{b_{k}} - v^{w_0}_{b_{k}} \Delta t
+ \frac{1}{2}g \Delta t^2 
\right) - \Delta p_{b_k}^{b_{k+1}}\\
&r_v = {q^{w_0}_{b_{k}}}^{-1} \otimes 
\left( v^{w_0}_{b_{k+1}} - v^{w_0}_{b_{k}}
+ g \Delta t
\right) - \Delta v_{b_k}^{b_{k+1}}\\
&r_{b_\omega} = b_{\omega_{k+1}} - b_{\omega_{k}}\\
&
r_{b_a} = b_{a_{k+1}} - b_{a_{k}}
    \end{split}
\end{align}

where
$\Delta q_{b_k}^{b_{k+1}}, \Delta p_{b_k}^{b_{k+1}}, \Delta v_{b_k}^{b_{k+1}}$ are the pre-integration of rotation, position and velocity between two consecutive frames. 

The residual of velocity is
\begin{equation}
    r(z^{g_{v_k}}, \chi ) = q_{w_0}^w \otimes v_{b_k}^{w_0} - v^w_{k}
\end{equation}
where
$q_{w_0}^w$ is the rotation error of the world frame estimated in the initialization step. We will optimize it using a large scale pose graph in the next section.
$v^w_{k}$ is the velocity of GNSS/RTK in the world frame. This term plays an important role to restrain the scale divergence of VI SLAM, and also help correct the course of motion.


\subsection{Large scale pose graph}

The step of pose graph integrates the poses estimated in multiple sliding windows, as well as GNSS positions and courses in a large scale. Furthermore, it also performs online correction for extrinsic parameters between GNSS antenna and IMU. The pose graph is performed in 4DOF since GNSS measurements in z-axis are ambiguous which may deteriorate the estimation of roll and pitch. Note that the IMU measurements have provided observable roll and pitch angles in BA process, we estimate only the yaw angle which gains the observability from GNSS measures.

The overall state variables are

\begin{equation}
   \begin{split}
       &\chi = \left[\zeta_0, \zeta_1, \zeta_2 ... \zeta_n, q_{w_0}^w, t^g_b \right]
      \\
       &\zeta_k = \left[ p^w_{b_k}, \phi^w_{b_k} \right]
       \\
       &\chi^* = \arg \min_{\chi} 
\left\{
\sum_{k=0}^{n} \left \| r(p^{b_{k+1}}_{b_k}, \chi ) \right \|^2_{\Omega_{p_k}} 
+\right.
\\
&
\sum_{k=0}^{n} \left \| r(\phi^{b_{k+1}}_{b_k}, \chi ) \right \|^2_{\Omega_{\phi_k}} 
+
\sum_{k=0}^{n} \left \| r(\phi^{b_{w}}_{b_k}, \chi ) \right \|^2_{\Omega_{\phi_{v_k}}}
+\\
&\left.
\sum_{k=0}^{n}\rho (\left \| r(\phi^{w}_g, \chi ) \right \|^2_{\Omega_{\phi_{w_k}}}) 
+
\sum_{k=0}^{n} \rho (\left \| r(p^{w}_g, \chi ) \right \|^2_{\Omega_{p_{w_k}}})
\right\}
   \end{split}
\end{equation}

\begin{align}
    \begin{split}
        r(p^{b_{k+1}}_{b_k}, \chi ) &= 
q(\phi^w_{b_k},\theta^w_{b_k},\gamma^w_{b_k})^{-1} \otimes 
(p^w_{b_{k+1}} - p^w_{b_k}) - \\
&{q^{w_0}_{b_k}}^{-1} \otimes (p^{w_0}_{b_{k+1}} - p^{w_0}_{b_k})\\
 r(\phi^{b_{k+1}}_{b_k}, \chi ) &= 
\{q(\phi^w_{b_k},\theta^w_{b_k},\gamma^w_{b_k})^{-1} \otimes 
q(\phi^w_{b_{k+1}},\theta^w_{b_{k+1}},\gamma^w_{b_{k+1}})\}^{-1}\\
&\otimes 
  \{{q^{w_0}_{b_k}}^{-1} \otimes 
{q^{w_0}_{b_{k+1}}}\} \\
 r(\phi^{b_{w}}_{b_k}, \chi) &= 
(q^w_{w_0} \otimes q^{w_0}_{b_k})^{-1} \otimes q(\phi^w_{b_k},\theta^w_{b_k},\gamma^w_{b_k})\\
 r(\phi^{w}_g, \chi ) &= 
\phi^w_{b_k} - \phi^w_{g_k}\\
 r(p^{w}_g, \chi ) &= 
p^w_{b_k} + q(\phi^w_{b_k},\theta^w_{b_k},\gamma^w_{b_k}) \otimes
t^b_g - p^w_{g_k}
    \end{split}
\end{align}

where,
$p^w_{b_k} $ is the position and
$\phi^w_{b_k}, \theta^w_{b_k}, \gamma^w_{b_k}$ are the yaw, pitch, roll angles of IMU in the world frame.
$p^{w_0}_{b_k}, q^{w_0}_{b_k}$
are the IMU poses in the initial world frame estimated in the BA process.
$q_{w_0}^w$ 
is the rotation error between the initial estimated world frame and real world frame.
$p^w_{g_k}, \phi^w_{g_k}$
are the position and course measurements of GNSS.
$r(p^{b_{k+1}}_{b_k}, \chi )$,$r(\phi^{b_{k+1}}_{b_k}, \chi )$,$r(\phi^{b_{w}}_{b_k}, \chi )$
are the prior residuals and $\Omega_{p_k},\Omega_{\phi_k}$, $\Omega_{\phi_{v_k}}$ the information matrixes for BA estimated relative translation, rotation and absolute yaw angle.
$r(\phi^{w}_g, \chi )$, $r(p^{w}_g, \chi )$
are the residuals and $\Omega_{p_{w_k}},\Omega_{\phi_{w_k}}$ the information matrixes for GNSS measured course and position.
$\rho(x)$is the huber loss function. We keep a large-scale sliding window to fuse global GNSS positions and courses. This plays an important role to avoid global position and yaw drift. Furthermore, outer-layer estimator run at a low frequency to reduce computational cost.

\section{EXPERIMENTAL RESULTS}
\label{experiment}
The proposed framework is evaluated on two public datasets.

    1. The famous indoor dataset EuRoc \cite{burri2016euroc}: the performance of the whole framework, especially the initialization is evaluated on each of 11 sequences, in comparison with other state-of-the-art methods.
    
    2. The outdoor dataset Kaist \cite{jeong2019complex}: We validate the long-term  practicability to outdoor challanges, e.g., the GNSS degradation, sparsity of feature points, low excitation of IMU, etc.

In addition, we also apply the proposed method on self-developed intelligent vehicle and conduct large-scale tests.
In each test, we estimate the camera pose fusing data from monocular camera, IMU and global positions.
All tests are run on an Intel Xeon(R) W-2125 CPU at 3.0GHz with 32 GB memeory.

\subsection{Experiment on Euroc}

We ran the experiments using only images from the left camera at 20Hz and IMU measurements at 200Hz. We add Gaussian noise to the ground-truth measurements to simulate global positions(refer to \cite{2020Tightly}). The standard deviation of the noise in each direction is  0.2m. The frequency of the signal is the same as the camera frequency.
We conducted a total of 11 sequences of the test. 5 sequences in the office
are labeled as MH\_, and others in the factory scene as V\_.


\subsubsection{Initialization performance}

As shown in Table \ref{initial_result}, the initialization performance is evaluated in terms of the scale error and the time cost.
On average, the scale errors of the proposed algorithm before and after BA are 22.82\% and 8.13\% respectively, which are significantly better than 52.15\% and 25.90\% of VINS-Mono. 
This is because we introduce the velocity of GNSS 
and use a tightly-coupled framework to efficiently and accurately correct the scale. 
The time consumption of calculating the initial value $t_{\text{Init}}$ and BA  $t_\text{BA}$ is close to that of VINS-Mono.
Note that VINS-Mono only performs visual-inertial  initialization, the proposed method additionally complete the  world frame alignment and therefore requires longer time $t_{\text{Tot}}$ to complete the overall initialization process.

\begingroup
\begin{table*}[t]
\scriptsize
\centering
\caption {Initialization Comparation: VINS-Mono with the proposed method}
\label{initial_result}
\setlength\tabcolsep{2.0pt}

\begin{tabular}{|c||cc|ccc|cc|ccc|}
\hline
  
 {} & \multicolumn{5}{c|}{VINS-Mono Initialization  \cite{qin2018vins} }  & \multicolumn{5}{c|}{ The proposed method}   \\ 
\cline{2-6} \cline{7-11} 
{} &  \multicolumn{2}{c|}{scale error (\%)} & & & & \multicolumn{2}{c|}{scale error (\%)} & & & \\
 \cline{2-3} \cline{7-8} 

      \begin{tabular}{@{}c@{}} Seq. \\ Name \end{tabular} & VI Align & \begin{tabular}{@{}c@{}} VI Align. \\ + BA  \end{tabular} & \begin{tabular}{@{}c@{}} $t_{Init}$ \\ (s) \end{tabular}  &
      \begin{tabular}{@{}c@{}} $t_{BA}$ \\ (s) \end{tabular} & \begin{tabular}{@{}c@{}} $t_{Tot}$ \\ (s) \end{tabular} &  \begin{tabular}{@{}c@{}} VI Align. \\ +GNSS \end{tabular}  & \begin{tabular}{@{}c@{}} VI + GNSS \\ + BA \end{tabular} & \begin{tabular}{@{}c@{}} $t_{Init}$ \\ (s) \end{tabular} & \begin{tabular}{@{}c@{}} $t_{BA}$ \\ (s) \end{tabular} & \begin{tabular}{@{}c@{}} $t_{Tot}$ \\ (s) \end{tabular} \\     \hline \\ \hline 
        
V1\_01 & 62.23 & 51.65 & 0.10 & 0.15 & \textbf{1.28} & \textbf{33.78} & \textbf{10.25} & \textbf{0.08} & \textbf{0.13} & 3.63 \\
V1\_02 & 65.65 & 36.83 & 0.10 & 0.14 & 1.45 & \textbf{26.83} & \textbf{7.23} & \textbf{0.09} & \textbf{0.13} & \textbf{1.14} \\
V1\_03 & 72.45 & 49.15 & 0.12 & 0.14 & \textbf{3.17} & \textbf{30.31} & \textbf{12.84} & \textbf{0.06} & \textbf{0.08} & 8.40 \\

V2\_01 & 45.21 & 33.26 & 0.15 & 0.21 & \textbf{1.09} & \textbf{22.77} & \textbf{7.07} & \textbf{0.05} & \textbf{0.09} & 1.54 \\
V2\_02 & 44.14 & 13.84 & 0.11 & 0.14 & \textbf{1.13} & \textbf{12.85} & \textbf{4.35} & \textbf{0.06} & \textbf{0.10} & 1.27 \\
V2\_03 & 70.36 & 27.75 & 0.12 & 0.15 & \textbf{2.80} & \textbf{30.14} & \textbf{4.34} & \textbf{0.07} & \textbf{0.09} & 6.34 \\
\hline 

MH\_01 & 25.20 & 13.34 & 0.13 & 0.19 & \textbf{2.49} & \textbf{10.01} & \textbf{5.26} & \textbf{0.06} & \textbf{0.10} & 2.52 \\
MH\_02 & \textbf{29.72} & 12.39 & 0.11 & 0.17 & \textbf{1.56} & 33.52 & \textbf{5.29} & \textbf{0.08} & \textbf{0.11} & 1.59\\
MH\_03 & 53.39 & 14.99 & 0.10 & 0.15 & 1.86 & \textbf{17.47} & \textbf{3.49} & \textbf{0.08} & \textbf{0.11} & \textbf{1.64} \\
MH\_04 & 49.22 & \textbf{10.93} & 0.11 & 0.16 & \textbf{1.18} & \textbf{10.74} & 14.65 & \textbf{0.05} & \textbf{0.08} & 2.03 \\
MH\_05 & 57.09 & 20.80 & 0.11 & 0.15 & 1.33 & \textbf{22.55} & \textbf{14.67} & \textbf{0.06} & \textbf{0.09} & \textbf{1.06} \\
\hline \\ \hline
Mean Values	& 52.15 & 25.90 & 0.11 & 0.13 & \textbf{1.76} & \textbf{22.82} & \textbf{8.13} & \textbf{0.07} & \textbf{0.10} & 2.83\\
\hline

\end{tabular}
\end{table*}
\endgroup

\subsubsection{Localization results}

\begin{figure}
    \centering
    \includegraphics[width = 0.42 \textwidth]{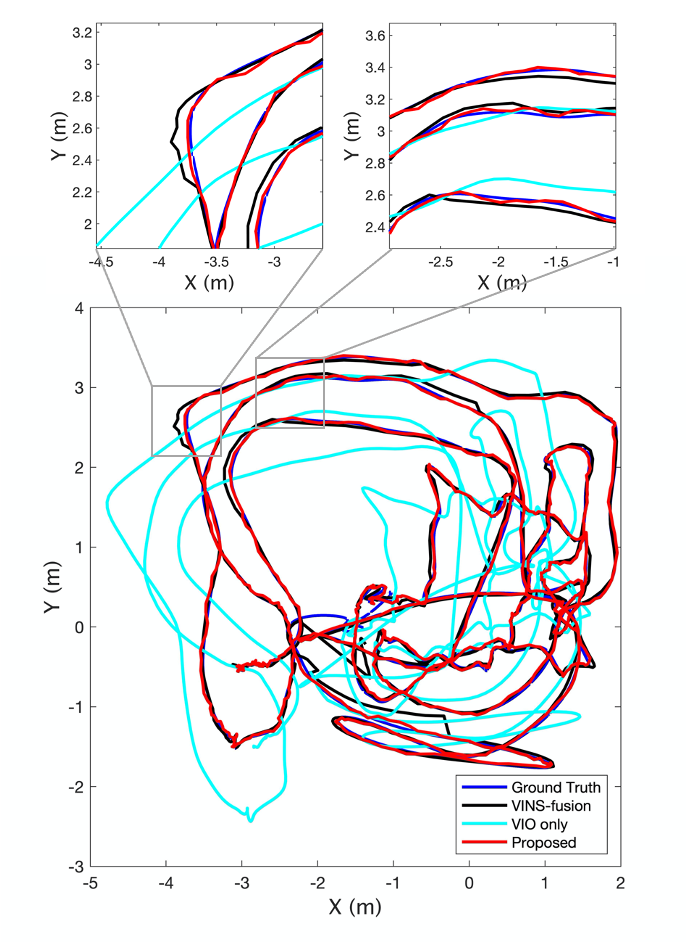}
    \caption{Localization result on EuRoC V2\_03}
    \label{eurocv203}
\end{figure}

In Figure \ref{eurocv203}, we take sequence V2\_03 as an example to show intuitively the  localization results.
It can be clearly seen that VINS-Mono trajectory diverges badly from the ground truth due to the accumulated position error and scale drift. VINS-Fusion \cite{qin2019general} performs significantly better thanks to the fusion of GNSS to eliminate accumulated errors. However, our proposed method provides the best performance. The improvement derives from more accurate initialization, as well as efficient usage of global measurements in the multi-layer estimator.

Table \ref{tab_euroc}
shows comparative results of each sequence in terms of the average translation error.
As can be seen, methods fusing GNSS are significantly more accurate than the VI only method.
Among the VI-GNSS fusion methods, the tightly-coupled one\footnote{Note that the result of \cite{2020Tightly} is directly quoted from the paper since the algorithm is not open-sourced.
We listed the result of N = 1 which is consistent with our setup.} has better performance than the loosely-coupled one (i.e., VINS-Fusion) due to the enhanced optimization from global measurements. However, it can also be sensitive to measurement noises. Our proposed method overcomes the problem using a multi-layer framework and achieves the best performance.

\begingroup
\begin{table}[ht]
\scriptsize
\centering
\caption {Localization results on the EuRoc dataset}
\label{tab_euroc}
\setlength\tabcolsep{2.0pt}

\begin{tabular}{|c||c|c|c|c|}
\hline

 {} & \multicolumn{4}{c|}{Translation mean error (m) }   \\ 
\cline{2-5}

 {} & VI only & \multicolumn{3}{c|}{VI-GNSS }   \\ 
\cline{2-5} 

\begin{tabular}{@{}c@{}} Seq. \\ Name \end{tabular}
& \begin{tabular}{@{}c@{}} VINS-Mono \end{tabular}& 
\begin{tabular}{@{}c@{}} VINS-\\Fusion \cite{qin2019general}   \end{tabular} &
\begin{tabular}{@{}c@{}} Tightly-\\coupled \cite{2020Tightly} \end{tabular} &
\begin{tabular}{@{}c@{}} Proposed \end{tabular}
\\
 \hline \hline 
     MH\_01 & 0.216 & 0.178 & \textbf{0.031} & 0.036 \\
    MH\_02 & 0.236 & 0.060 & 0.036 & \textbf{0.026}  \\
     MH\_03 & 0.273 & 0.090 & 0.048 & \textbf{0.044}  \\
      MH\_04 & 0.191 & 0.044 & 0.068 & \textbf{0.026}  \\
      MH\_05 & 0.551 & 0.083 & 0.056 & \textbf{0.040}  \\
        V1\_01 & 0.139 & 0.048 & 0.041 & \textbf{0.035}  \\
        V1\_02 & 0.112 & 0.055 & 0.048 & \textbf{0.026}  \\
        V1\_03 & 0.259 & 0.034 & 0.068 & \textbf{0.025}  \\
        V2\_01 & 0.104 & 0.100 & 0.038 & \textbf{0.022}  \\
        V2\_02 & 0.152 & 0.048 & 0.046 & \textbf{0.033}  \\
        V2\_03 & 0.385 & 0.343 & 0.098 & \textbf{0.029}  \\ 
 \hline
\end{tabular}
\end{table}
\endgroup

\subsection{Outdoor experiments}


We conduct experiments 
on the large-scale dataset Kaist \cite{jeong2019complex} and compare the result with VINS-Fusion. 
The camera is Pointgrey Flea3 with the resolution of 1600$\times$1200 and the acquisition frequency of 10Hz. The IMU is Xsens’s MTi-300, with frequency of 100Hz. The RTK-GNSS is SOKKIA’s GRX2, with frequency of 1 Hz. Since the GNSS data has a low frequency and is not synchronized with other sensors, we linearly interpolate the GNSS data at each image frame.
The dataset also parovide ground-truth. It should be noted that, as the authors  mentioned \cite{jeong2019complex}, due to the very complicated experimental scene, the ground-truth also has noise.

\subsubsection{Localization results}
The first three sequences contain typical urban scenes, and are labeled as urban\_.
The last three sequences are highway scenes, and are labeled as highway\_. The total length for urban\_ and highway\_ are 29.99km and 8.8km respectively. 






\begin{figure}
    \centering
    \includegraphics[width = 0.40 \textwidth]{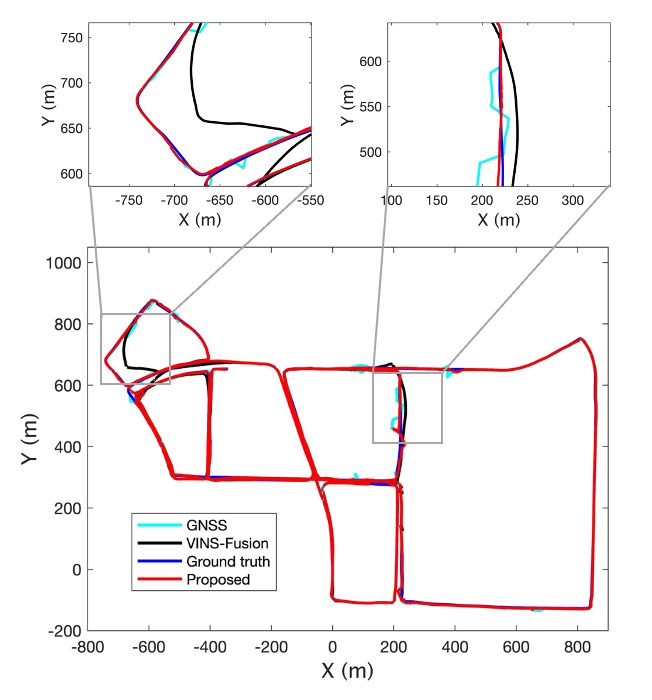}
    \caption{Localization result on urban\_1}
    \label{kaist39}
\end{figure}

Figure \ref{kaist39} shows comparatively the trajectories of urban\_1.
Since VINS-Fusion does not estimate the local-global transformation in the real-time, the trajectory is produced by offline optimization. Note that the trajectory of our proposed method is generated in the online mode.

It is obvious from Figure \ref{kaist39} that
VINS-Fusion improves the trajectory smoothness in urban canyons where GNSS drift occurs. However, continuous GNSS errors (refer to the upper figures in Figure \ref{kaist39}) will reveal the limitations of accumulated scale drift in VI fusion and fragile local-global alignment in the VI-GNSS loosely-coupled framework.
Our proposed algorithm eliminates scale drift in the inner-layer using global measurement and estimates the local-global transformation more accurately in the outer-layer. The improvement ensures enhanced robustness to long-term GNSS degradation.

\begin{figure}
    \centering
    \includegraphics[width = 0.45 \textwidth]{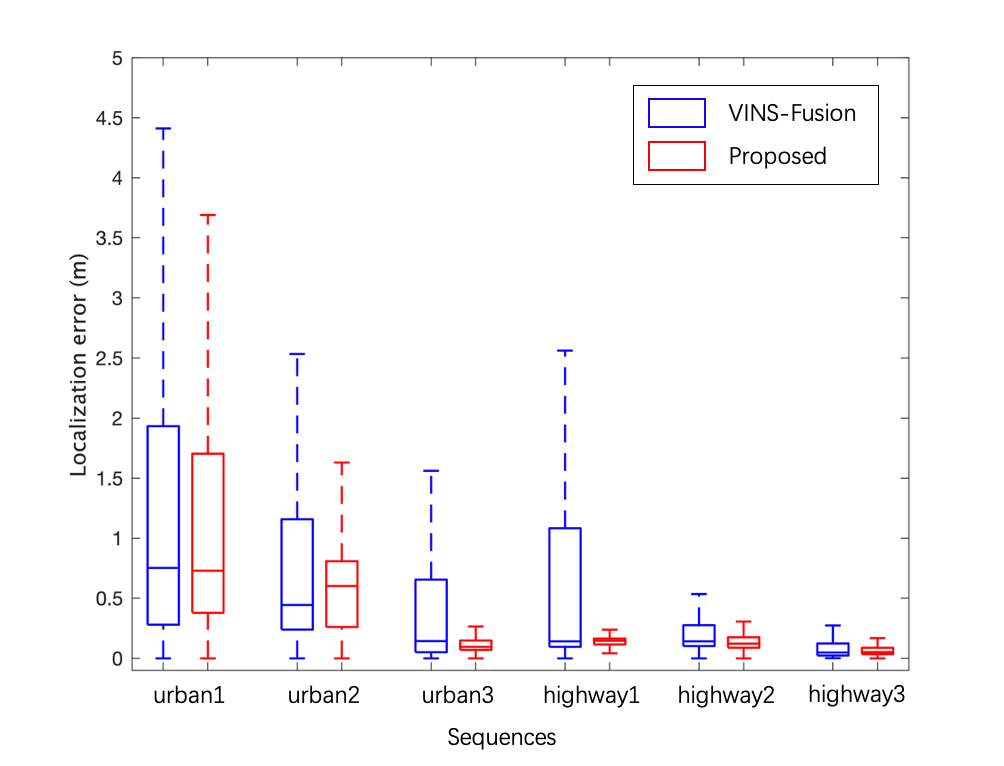}
    \caption{Localization error distribution of Kaist dataset}
    \label{kaist_box}
\end{figure}

Figure \ref{kaist_box} shows the localization error statistics with box plots of all 
sequences. Our proposed method significantly outperforms VINS-Fusion in both urban and highway scenarios.
In urban scenarios, the improvement derives from the enhanced robustness to GNSS drift, as we mentioned above. The highway scenario faces challenges in two aspects:
 1) the visual features are sparse and are far away from the camera; 2) the IMU excitation is weak due to the limited speed variation. The problems therefore cause accumulated scale drift. Our proposed method fuses the GNSS velocity in the inner-layer BA to continuously correct the metric scale. As can be seen, tests on all three highway sequences achieve very high localization accuracy.



\subsection{Application on intelligent vehicle}

We apply the proposed framework to a self-developed intelligent vehicle.
The localization platform is equipped with a monocular rolling shutter camera which costs as low as 10\$. The resolution is 1920 $\times$ 1080.  The 
IMU module costs 2\$, and its data frequency is 412Hz. The GNSS module costs 10\$ working under the RTK mode at 10Hz.
The sensors are synchronized using Pulse Per Second (PPS) signal.
The ground truth is obtained by post-process of a high-proformance GNSS-inertial integration device  Trimble AP60.



We conducted tests on expressways and urban roads in Chaoyang District, Beijing. The total length is 36.04km.
As shown in Table 
\ref{tab_asample},
the localization error reduced by 63\% compared with VINS-Fusion. 


\begingroup
\begin{table}[ht]
\scriptsize
\centering
\caption {Localization error on self-developed experimental platform}
\label{tab_asample}
\setlength\tabcolsep{2.0pt}

\begin{tabular}{|c||c|c|c|c|}
\hline
  
 {} & \multicolumn{4}{c|}{Translation error (m) }   \\ 
\cline{2-5} 

Method. 
& 
\begin{tabular}{@{}c@{}} Mean \end{tabular}
& 
\begin{tabular}{@{}c@{}} Max \end{tabular}&
90\% & 
95\%
\\
 \hline 
     GNSS& 0.52 & 32.51 & 1.34 & 1.51 \\
    VINS-Fusion& 0.42 & 14.43 & 0.78 & 1.53 \\
     Proposed & \textbf{0.19} & \textbf{7.51} & \textbf{0.43} & \textbf{0.86}  \\
 \hline
\end{tabular}
\end{table}
\endgroup

\section{CONCLUSIONS}

This paper proposes a high-precision localization method based on vision-inertial and GNSS fusion.
A novel hierarchical framework is developed to fuse various sensor information efficiently in multiple layers to achieve accurate and reliable real-time positioning. 
In order to quickly obtain an accurate initial values, we also propose a numerical solution aided MAP initialization method.
Tests on indoor and outdoor datasets, as well as on our self-developed platform show that the algorithm in this paper can achieve more accurate real-time localization compared to existing algorithms in large-scale scenarios.




\bibliographystyle{unsrt}
\bibliography{root}

\end{document}